\documentclass[final]{cvpr}

\usepackage{times}
\usepackage{epsfig}
\usepackage{graphicx}
\usepackage{amsmath}
\usepackage{amssymb}
\usepackage{xspace}

\usepackage[pagebackref=true,breaklinks=true,colorlinks,bookmarks=false]{hyperref}

\usepackage{graphicx}
\usepackage{comment}
\usepackage{xcolor}
\usepackage{mathtools}
\usepackage{capt-of}
\usepackage{booktabs}
\usepackage{stfloats}
\newcommand{\OURS}[1]{ComPoseNet}
\newcommand{\OURSSEG}[1]{ComPoseNet w/o compl.}
\newcommand{\FTFGT}[1]{F2F-GT}
\newcommand{\FTFMASKRCNN}[1]{F2F-MaskRCNN}
\newcommand{\MASKFUSION}[1]{MaskFusion}
\newcommand{\MIDFUSION}[1]{MID-Fusion}
\newcommand{\DYNSYNTH}{\textsc{DynSynth}\xspace}

\def\cvprPaperID{4307} % *** Enter the CVPR Paper ID here
\def\confYear{CVPR 2021}

\begin{document}

%%%%%%%%% TITLE
\title{Seeing Behind Objects for 3D Multi-Object Tracking in RGB-D Sequences}

\newcommand{\footremember}[2]{%
    \footnote{#2}
     \newcounter{#1}
     \setcounter{#1}{\value{footnote}}%
 }

 \author{
 Norman Müller$^{1}$ \qquad   Yu-Shiang Wong$^{2}$ \qquad  Niloy J. Mitra$^{2,3}$ \qquad  Angela Dai$^{1}$  \qquad       Matthias Nie{\ss}ner$^{1}$ \\
 \qquad \\
 $^{1}$Technical University of Munich \qquad $^{2}$University College London \qquad  $^{3}$Adobe Research \\
 }

\twocolumn[{%
\renewcommand\twocolumn[1][]{#1}%
\maketitle
\centering
\includegraphics[width=\linewidth]{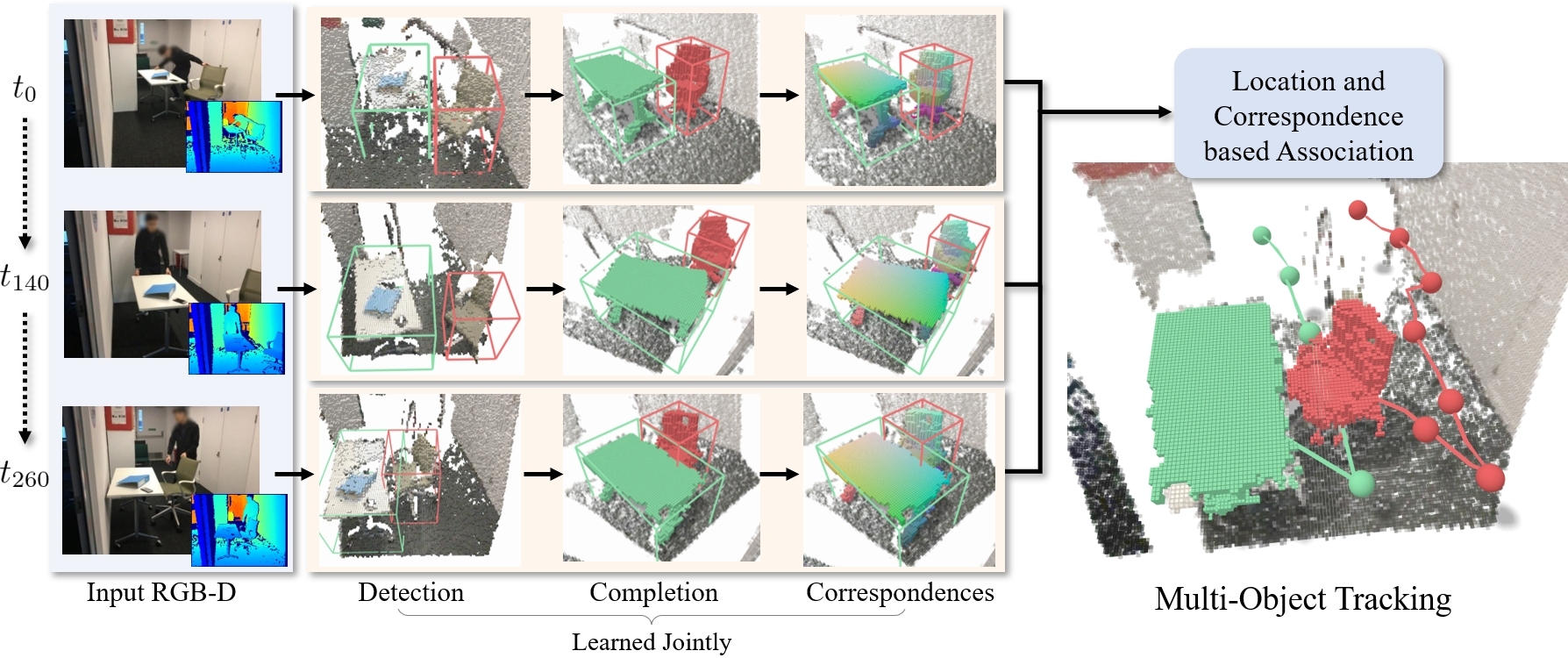}
\captionof{figure}{Our method learns to see behind objects in RGB-D sequences in order to achieve robust dynamic object tracking; we predict the complete underlying geometry of each object beyond the observed view, which enables finding correspondences which can more reliably persist over time, under various view changes and object motion.
        From an input RGB-D frame, we first perform 3D object detection, then jointly infer for each object its complete geometry and dense correspondence mapping to its canonical space. 
        These correspondences on the predicted complete object geometry help to provide robust multi-object tracking over time.}
\label{fig:teaser}
\vspace*{10px}
}]

\begin{abstract}
Multi-object tracking from RGB-D video sequences is a challenging problem due to the combination of changing viewpoints, motion, and occlusions over time.
We observe that having the complete geometry of objects aids in their tracking, and thus propose to jointly infer the complete geometry of objects as well as track them, for rigidly moving objects over time.
Our key insight is that inferring the complete geometry of the objects significantly helps in tracking.
By hallucinating unseen regions of objects, we can obtain additional correspondences between the same instance, thus providing robust tracking even under strong change of appearance.
From a sequence of RGB-D frames, we detect objects in each frame and learn to predict their complete object geometry as well as a dense correspondence mapping into a canonical space.
This allows us to derive 6DoF poses for the objects in each frame, along with their correspondence between frames, providing robust object tracking across the RGB-D sequence.
Experiments on both synthetic and real-world RGB-D data demonstrate that we achieve state-of-the-art performance on dynamic object tracking.
Furthermore, we show that our object completion significantly helps tracking, providing an improvement of $6.5\%$ in mean MOTA.

\end{abstract}

\section{Introduction}

Understanding how objects move over time is fundamental towards higher-level perception of real-world environments, with applications ranging from mixed reality to robotic perception.
In the context of static scenes, significant progress has been made in RGB-D tracking and reconstruction \cite{newcombe2011kinectfusion,izadi2011kinectfusion,niessner2013hashing,whelan2015elasticfusion,choi2015robust,dai2017bundlefusion}; however, the assumption of a static environment significantly limits applicability to real-world environments which are often dynamic, with objects moving over time. 
In the case of scenes where a number of objects might be rigidly moving, robust tracking remains a significant challenge, as views and occlusion patterns of the objects can change appreciably over time.

Several approaches have been developed to address the problem of dynamic object tracking in RGB-D sequences by detecting objects and then finding correspondences between frames~\cite{runz2017co, runz2018maskfusion, xu2019mid}.
While results have shown notable promise, these methods only consider the observed geometry of the objects, and so tracking objects under faster object or camera motion can result in insufficient overlap of observed geometry to find reliable correspondences, resulting in tracking failure.

To address these challenges, we observe that humans can effectively track objects by leveraging prior knowledge of the underlying object geometry, which helps to constrain the problem even under notable view changes or significant occlusions.
Thus, our key idea is to learn to `see behind objects' by \emph{hallucinating the complete object geometry in order to aid object tracking}.
We learn to jointly infer for each object its complete geometry as well dense tracking correspondences, providing 6DoF poses for the objects for each frame.

From an RGB-D sequence, we formulate an end-to-end approach to detect objects, characterized by their 3D bounding boxes, then predict for each object its complete geometry as well as a dense correspondence mapping to its canonical space. 
We then leverage a differentiable pose optimization based on the predicted correspondences of the complete object geometry to provide the object poses per frame as well as their correspondence within the frames.

Our experiments show that our joint object completion and tracking provides notably improved performance over state of the art by 6.5\% in MOTA.
Additionally, our approach provides encouraging results for scenarios with challenging occlusions.
We believe this opens up significant potential for object-based understanding of real-world environments.

\section{Related Work}

\paragraph{RGB-D Reconstruction of Static Scenes}

Scanning and reconstruction 3D surfaces of static environments has been widely studied~\cite{newcombe2011kinectfusion,izadi2011kinectfusion,choi2015robust,whelan2015elasticfusion,dai2017bundlefusion}, with state-of-the-art reconstruction approaches providing robust camera tracking of large scale scenes. 
While these methods show impressive performance, they rely on a core, underlying assumption of a static environment, whereas an understanding of object movement over time can provide a profound, object-based perception.

Various approaches have also been developed for static scene reconstruction to simultaneously reconstruct the scene while also segmenting the observed geometry into semantic instances \cite{tateno20162,salas2013slam++,mccormac2017semanticfusion,mccormac2018fusion++}.
Notably, Hou et al.~\cite{hou2020revealnet} propose to jointly detect objects as well as infer their complete geometry beyond the observed geometry, achieving improved instance segmentation performance; however, their method still focuses on static environments.
In contrast, our approach exploits learning the complete object geometry in order to object tracking in dynamic scenes.

\paragraph{RGB-D Object Tracking}
Several approaches have been proposed towards understanding dynamic environments by object tracking.
To achieve general non-rigid object tracking, research focuses on the single object scenario, typically leveraging as-rigid-as-possible registration~\cite{zollhofer2014real,newcombe2015dynamicfusion,innmann2016volumedeform,dou2016fusion4d,guo2017real,bozic2020deepdeform}.
For multiple object tracking, object rigidity is assumed, and objects are detected and then tracked over time.
In the context of SLAM, SLAMMOT~\cite{wang2007simultaneous}, and CoSLAM~\cite{zou2012coslam} demonstrated detection and tracking of objects, operating with sparse reconstruction and tracking.
Co-Fusion~\cite{runz2017co}, MID-Fusion~\cite{xu2019mid}, and MaskFusion~\cite{runz2018maskfusion} demonstrated dense object tracking and reconstruction, with promising results for dynamic object tracking, but can still suffer noticeably from occlusions and view changes, as only observed geometry is considered.
Our approach not only reconstructs the observed geometry of each object, but infers missing regions that have not been seen, which is crucial to achieve robust object tracking under these challenging scenarios.

\section{Method Overview}

Our method takes as input an RGB-D sequence, and learns to detect object instances, and for each instance the per-frame 6DoF poses and dense correspondences within the frames. 
We then associate the predicted locations and correspondences to obtain object tracking over time.

Each RGB-D frame of the sequence is represented by a sparse grid $\mathcal{S}_i$ of surface voxels and a dense truncated signed distance field (TSDF) $\mathcal{D}_i$.

The TSDF for an RGB-D frame is obtained by back-projecting the observed depth values, following volumetric fusion~\cite{curless1996volumetric}.

As output, we characterize each detected object in every frame with a 3D occupancy mask representing its complete geometry along with a dense grid of correspondences to the object's canonical space, from which we compute the 6DoF pose.
We then use the complete correspondence prediction to associate objects across time steps, resulting in robust multi-object tracking over time.

From the input sparse surface grid, we detect objects by regressing their 3D object centers and extents, and cluster them into distinct bounding box proposals.

For each object proposal, we crop the TSDF volume using the respective bounding box, and use this information to predict the object's complete geometry as a dense occupancy grid as well as its normalized object coordinates mapping the object to its canonical space.

We can then solve for the object pose using a differentiable Procrustes analysis.

To perform multi-object tracking across the RGB-D sequence, we associate instances across the frames based on 3D bounding box overlap as well as the 3D intersection-over-union of the predicted complete canonical geometry.
Predicting the underlying geometric structure of each object enables our approach to maintain robustness under large camera pose changes or object movement, as we can associate the complete object geometry beyond the observed regions.
Thus, from our object detection and then completion, we are able to find more correspondences which can persist over the full sequence of frames, providing more overlap for an object between frames, and resulting in more robust object instance tracking.

\section{Joint Object Completion and Tracking}

\begin{figure*}
	\centering
    \includegraphics[width=\linewidth]{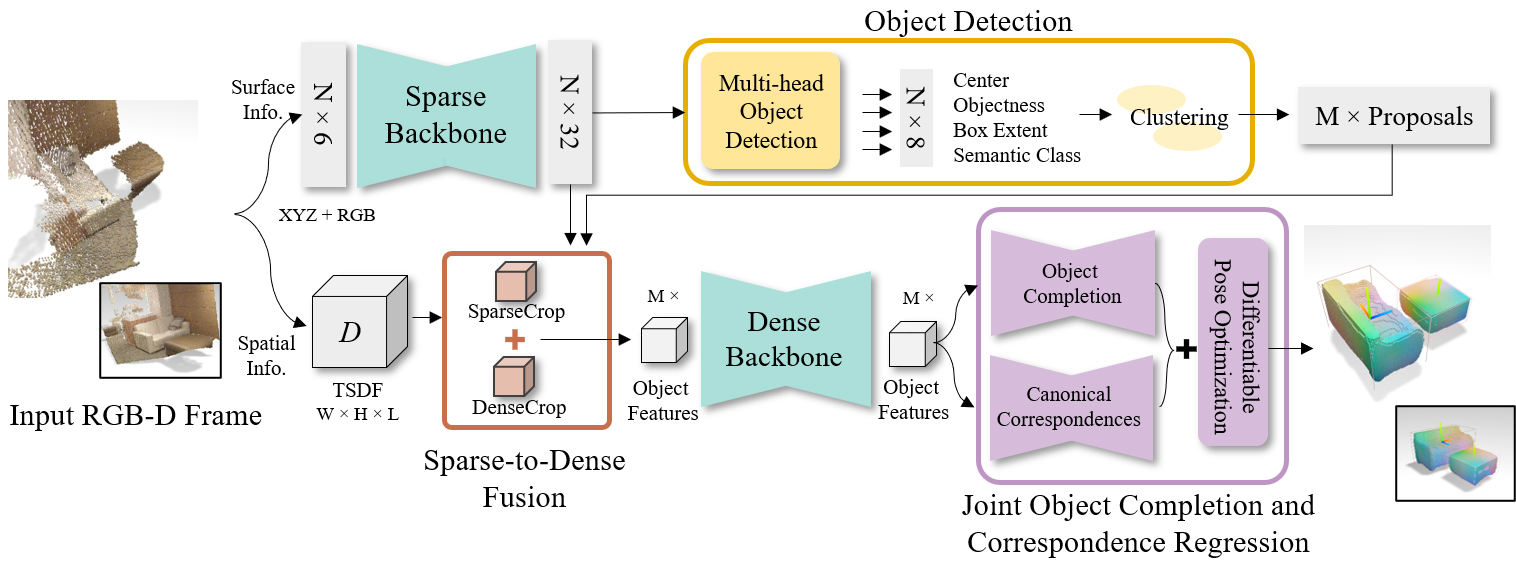}
	\caption{
	    Overview of our network architecture for joint object completion and tracking.
	    From a TSDF representation of an RGB-D frame, we employ a backbone of sparse 3D convolutions to extract features.
	    We then detect objects characterized by 3D bounding boxes, and predict for each object both the complete object geometry beyond the view observation as well as dense correspondences a canonical space; the correspondences on the complete geometry then inform a differentiable pose optimization to produce object pose estimates and within-frame dense correspondences.
	    By predicting correspondences not only in observed regions but also unobserved areas, we can provide strong correspondence overlap under strong object or camera motion, enabling robust dynamic object tracking.
    }
	\label{fig:architecture}
\end{figure*}

From an RGB-D sequence, we first detect objects in each frame, then infer the complete geometry of each object along with its dense correspondences to its canonical space, followed by a differentiable pose optimization.

An overview of our network architecture for joint object completion and correspondence regression is shown in Figure~\ref{fig:architecture}.
From an object detection backbone, we simultaneously predict an object's complete geometry and dense correspondences, which informs its pose optimization. For a detailed architecture specification, we refer to the supplemental.

\subsection{Object Detection}
We first detect objects from the sparse surface grid $\mathcal{S}$ for each RGB-D frame by predicting their object bounding boxes.
We extract features from the sparse surface grid using a series of sparse convolutions~\cite{graham20183d,choy20194d} structured in encoder-decoder fashion, with features spatially bottlenecked to $1/16$ of the original spatial resolution, and the output of the final decoder layer equal to the original spatial resolution.
The feature map $F$ from the last decoder layer is passed as input to a multi-head object detection module.
The detection module predicts objectness, with each voxel $v$ predicting $O(v)$ as the score that $v$ is associated with an object, the 3D center location $C(v)$ of the object as a relative offset from $v$, and the 3D extents $D(v)$ of the object as well as the semantic class $S(v)$.
We then train using the following loss terms:
\begin{align*} 
    L_o &= BCE(O, O^t) \\
    L_c &= \begin{cases}
        \frac{1}{2}{(C - C^t)^2}                   & \text{for } |C - C^t| \le 0.5, \\
        |C - C^t| - \frac{1}{2}, & \text{otherwise} 
    \end{cases} \\
    L_d &= \begin{cases}
        \frac{1}{2}{(D-D^t)^2}                   & \text{for } |D - D^t| \le 0.5, \\
        |D - D^t| - \frac{1}{2}, & \text{otherwise,} 
    \end{cases} \\
    L_s &= CE(S,S^t)
\end{align*}
with $O^t$ denoting the target objectness as a binary mask of the target objects' geometry, and $C^t$, $D^t$ and $S^t$ the target object centers, extents and semantic class, respectively, defined within the mask of the target objects' geometry.

To obtain the final object proposals, we perform a mean-shift clustering (20 steps, with 8 voxel radius) on the predicted center coordinates of the voxels which produce a positive objectness score.
From the resulting instance clusters, we filter out small clusters of less than 50 elements. On the remaining clusters, we perform average pooling on the  bounding box extent predictions and majority voting on the highest scoring semantic classes for final object location, shape and semantic class prediction.

\paragraph{Sparse-to-Dense Fusion.}
For each detected object and its predicted box, we then crop the corresponding sparse features $f_k$ from $F$ as well as the dense TSDF grid $\mathcal{D}$. We map the sparse cropped features densely and add the matching TSDF values over the feature channels to obtain $f_k'$.
We can then leverage this feature to inform object completion and correspondence regression in both observed and un-observed space.

\subsection{Object Completion}

To predict the complete object geometry, we take the  sparse-dense fused feature $f_k'$  for an object $k$, which is then down-scaled by a factor of 2 using trilinear interpolation and passed through a series of dense 3D convolutions, structured in encoder-decoder fashion to obtain dense object features $f_k^o$. 
We then apply another series of dense 3D convolutional layers on  $f_k^o$ to predict the complete object geometry $m_k$ as a binary mask trained by binary cross entropy with the target occupancy grid.

\subsection{Object Correspondences}

We predict for each object a dense correspondence mapping $c_k$ to its canonical space, similar to the normalized object coordinate space of \cite{wang2019normalized}.
Using both $c_k$ and the object geometry $m_k$, we can perform a robust pose optimization under the correspondences.

The correspondences $c_k$ are predicted from the object feature map ${f_k^o}'$ by a series of dense 3D convolutions structured analogously to the object geometry completion, outputting a grid of 3D coordinates in the canonical space of the object.
We apply an $l_1$ loss to the $c_k$, evaluated only where target object geometry exists.

To obtain the object pose in the frame, we take the correspondences from $c_k$ where there is object geometry (using target geometry for training, and predicted geometry at test time), and optimize for the object rotation and scale under the correspondences using a differentiable Procrustes analysis.

We aim to find scale $c^*$, rotation $R^*$ and translation $t^*$ that bring together predicted object coordinates $P_o$ with their predicted canonical representation $P_n$:
\begin{equation}
c^*,R^*,t^* := \underset{c\in\mathbb{R}^+, R\in SO_3, t\in \mathbb{R}^3 }{argmin}{\|P_o -  (cR\cdot P_n +t)\|}.
\end{equation}
With means $\mu_i$ and variances $\sigma_i$ of $P_i$, $i\in \{o,n\}$, we perform a differentiable SVD of $(P_o - \mu_o)(P_n - \mu_n)^T = UDV^T$. According to \cite{umeyama1991}, with $S=diag(1,1,det(UV^T))$, we obtain the optima
\begin{equation}
        c^{*}=\frac{1}{\sigma_n}tr(DS),  R^{*} = USV^T \text{, and } t^*=\mu_o - c^*R^*\mu_n. 
\end{equation}

We employ a Frobenius norm loss on the estimated rotation matrix, an $\ell_1$ loss on the predicted scale, and an $\ell_2$ loss on the translation.

Since objects possessing symmetry can result in ambiguous target rotations, we take the minimum rotation error between the predicted rotation and the possible valid rotations based on the object symmetry.

\subsection{Object Tracking}
Finally, to achieve multi-object tracking over the full RGB-D sequence, we associate object proposals across time steps, based on location and canonical correspondences. 
Each detected object has a predicted bounding box and canonical object reconstruction, represented as a $64^3$ grid by mapping the dense correspondences in the predicted object geometry to canonical space.
To fuse detections over time into tracklets, we construct associations in a frame-by-frame fashion; we start with initial tracklets $T^i$ for each detected object in the first frame.

Then, for each frame, we compute pairwise distances between current tracklets $T^i$ and incoming proposals $D^j$ based on the 3D IoU of their bounding boxes. 
We employ the Hungarian algorithm~\cite{kuhn1955} to find the optimal assignment of proposals to tracklets, and  reject any matches with 3D IoU below 0.3.
Any new object detections with no matches form additional new tracklets.
The canonical object reconstruction for a tracklet is then updated as a running average of the canonical reconstructions for each object detection in that tracklet; we use a 4:1 weighting for the running mean for all our experiments.
After computing the tracklets and their canonical reconstructions from the frames in sequential order, we then aim to match any objects which might have not have been matched in the greedy sequential process (e.g., seen from a very different view, but able to match to the full reconstruction from many views).
For all tracklets and all non-assigned proposals, we compute pairwise distances using a 3D volumetric IoU of the canonical representations (binarized at threshold 0.5). 
We again compute the optimal assignment and reject a matching if this mask IoU is below 0.3.

We find that by matching objects based on their canonical correspondences, we observe higher matching accuracy, leading to robust object tracking (see Section~\ref{sec:results}).

\subsection{Training Details}
We train our joint object completion and correspondence regression on a single Nvidia GeForce RTX 2080, using an ADAM optimizer with learning rate 0.001 and weight decay of 1e-5.
We use a batch size of $2$, and up to $10$ proposals per input.
To provide initial stable detection results, we first train the object detection backbone for 100K iterations, and then introduce the object completion and correspondence prediction along with the differentiable pose optimization, training the full model end-to-end for another 250K iterations until convergence.
Full training takes approximately $72$ hours.

We weight the object center and extent loss, $L_c$ and $L_d$ by $0.1$, as they are evaluated in voxel units with have larger absolute value. After a warm-up phase of 100k iterations, where segmentation, detection and completion are trained individually, we weight the completion and correspondence loss by $4$, and the rotation, translation and scale loss by $0.2$, $0.1$,$0.1$, respectively, to bring the loss values into similar ranges.

\section{Results}
\label{sec:results}

\begin{table*}[tp]
\centering
\begin{tabular}{l|cccccccccc|c}
\hline
MOTA(\%)                   & bathtub       & bed          & bookshelf     & cabinet       & chair         & desk          & sink          & sofa          & table         & toilet        & seq. avg      \\ \hline
\MASKFUSION{}~\cite{runz2018maskfusion}                    & 27.7          & 76.4         & 25.4          & 24.4          & 25.3          & 33.8          & 39.2          & 5.7           & 45.8          & 27.7          & 17.2               \\
\MIDFUSION{}~\cite{xu2019mid}                     & \textbf{55.8 }         & \textbf{100 }         & \textbf{94.7} & 21.7          & 38.6          & 45.8          & 63.9          & 9.6           & 53.8          & 35.7          & 30.1          \\
\FTFMASKRCNN{}                   & 25.7          & \textbf{100 }         & 73.7          & 15.2             & 28.3          & \textbf{79.2}          & \textbf{73.2}          & \textbf{21.2 }         & 59.6          & 33.9          & 35.8          \\ \hline\hline
Ours (no corr., no compl. )  & 39.8          & 54.5         & 22.6          & 21.8          & 27.2          & 37.5          & 49.5          & 13.8          & 60.4          & 36.7          & 29.3          \\
Ours (no corr.) & 39.8          & 54.5         & 24.0          & 23.2          & 32.2          & 37.5          & 50.3          & 13.8          & 61.8          & 38.1          & 30.6          \\
Ours (no compl.)              & 24.9          & 45.5         & 50.0          & 26.1          & 42.3          & 66.4          & 63.3          & 18.0          & 63.2          & 38.0          & 35.6          \\
Ours                          & 24.9          & 45.5         & 50.1          & \textbf{26.1} & \textbf{51.8} & 66.4          & 63.3          & 17.3          & \textbf{67.4} & \textbf{49.0} & \textbf{42.3}
\end{tabular}
\vspace{0.1cm}
\caption{Evaluation of MOTA on \DYNSYNTH{}.
    Our approach to jointly predict complete object geometry along with tracking provides robust correspondences over the full object rather than only the observed regions, resulting in notably improved tracking in comparison to our approach without object completion (\emph{no compl.}), purely IoU based matching (\emph{no 
    corr.}) as well as state of the art. 
    }
	\label{tab:mota_class} 
\vspace{0.1cm}
\end{table*}

\begin{figure*}
	\centering
	\includegraphics[width=0.95\linewidth]{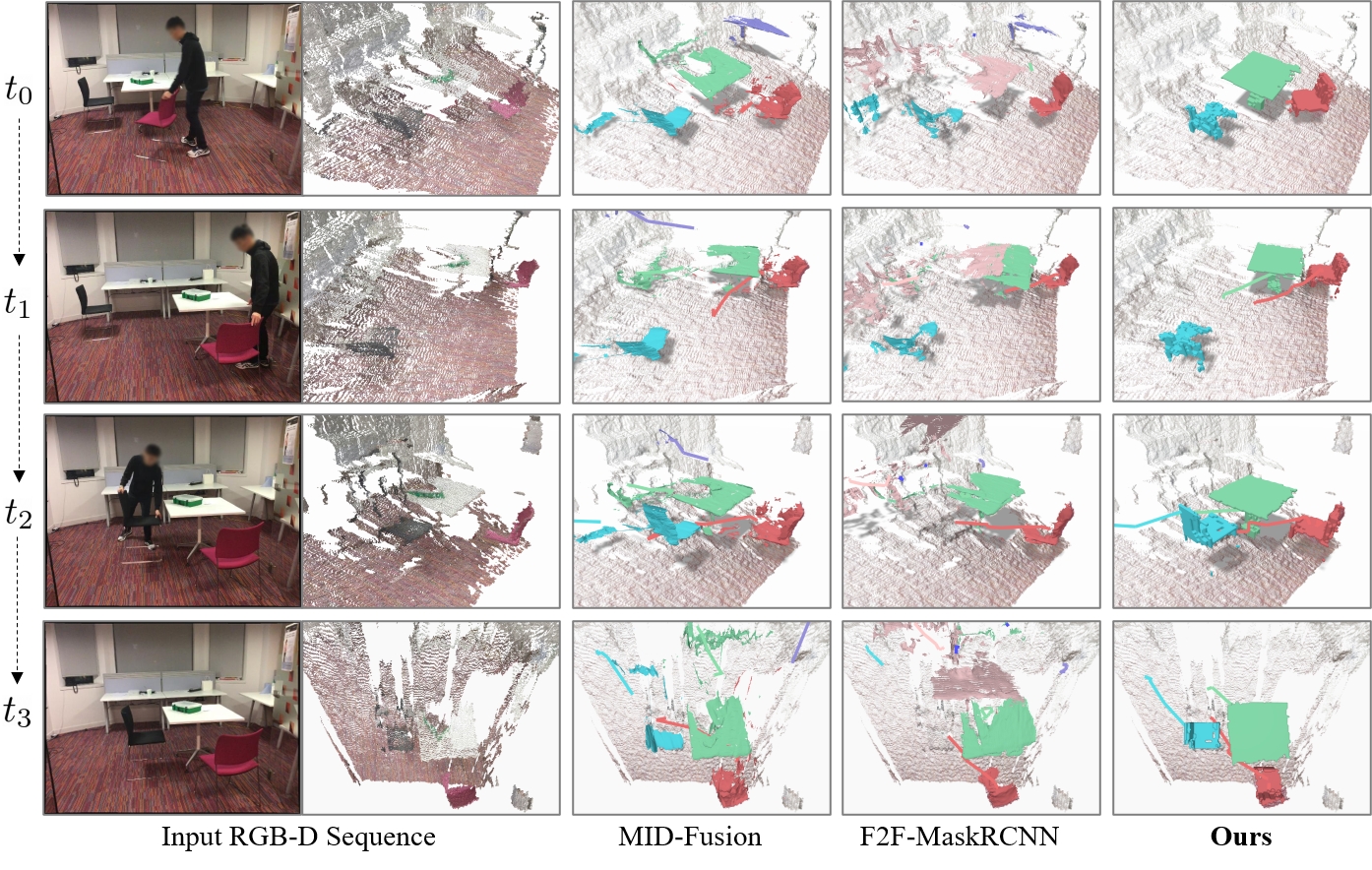}
	\includegraphics[width=0.95\linewidth]{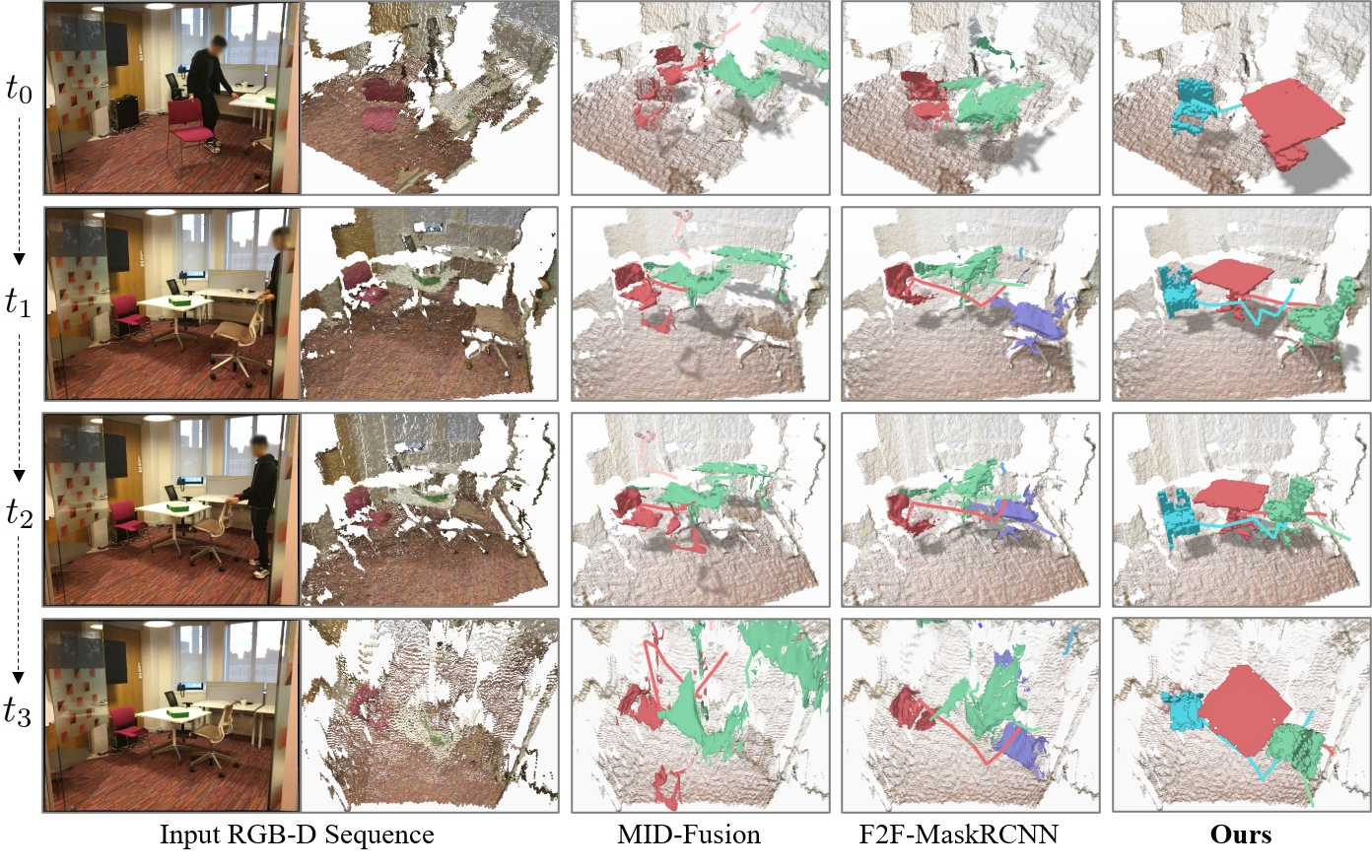}
	\caption{
	     Our joint object completion and tracking on real-world RGB-D sequences maintains consistent objects tracks and accurate object shapes over time. The colors and the line segments show the instance ID and the estimated trajectories, respectively.
    }
	\label{fig:comparison_real}
\end{figure*}

\begin{figure*}
	\centering
	\includegraphics[width=0.95\linewidth]{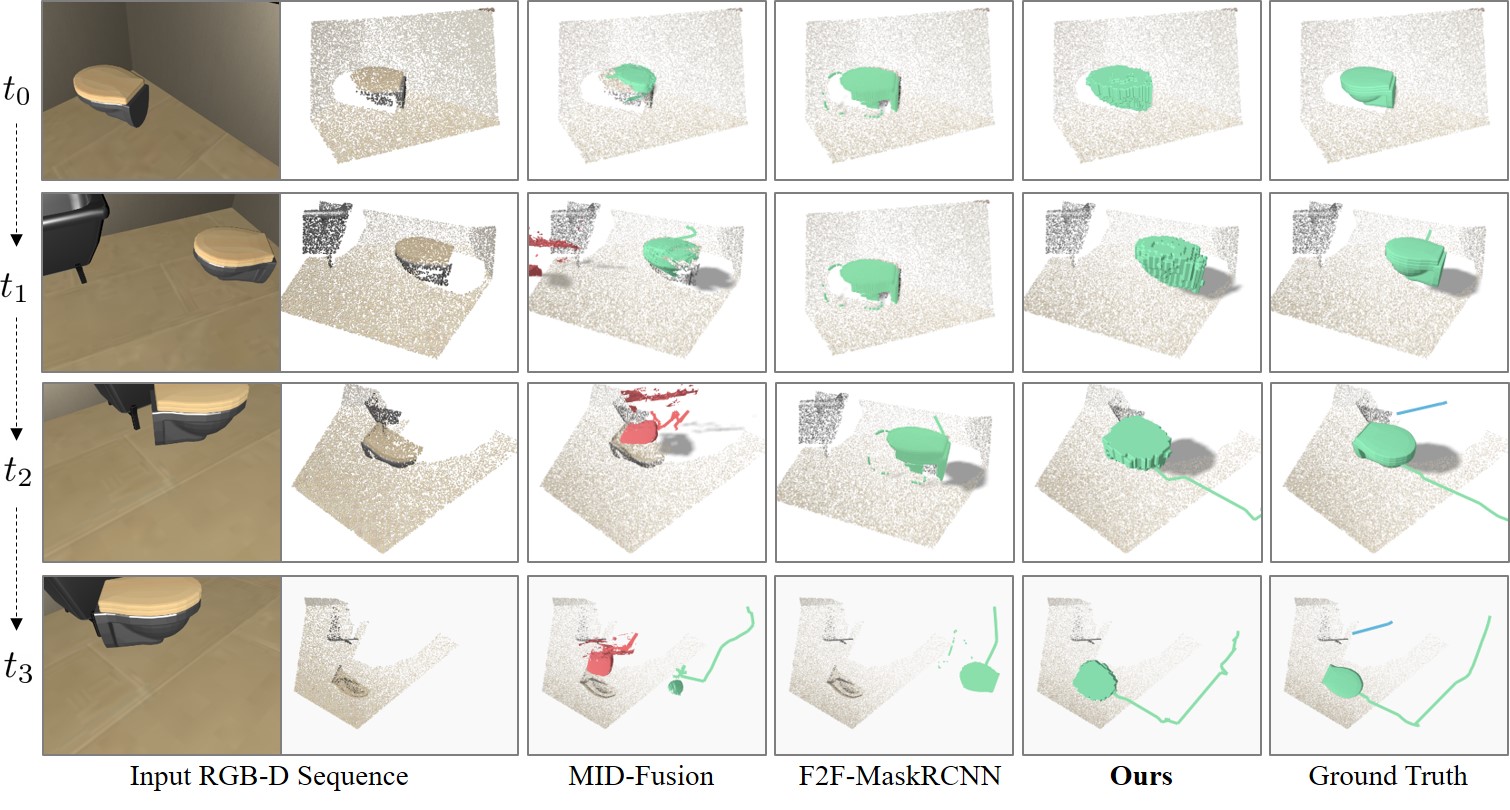}
	\includegraphics[width=0.95\linewidth]{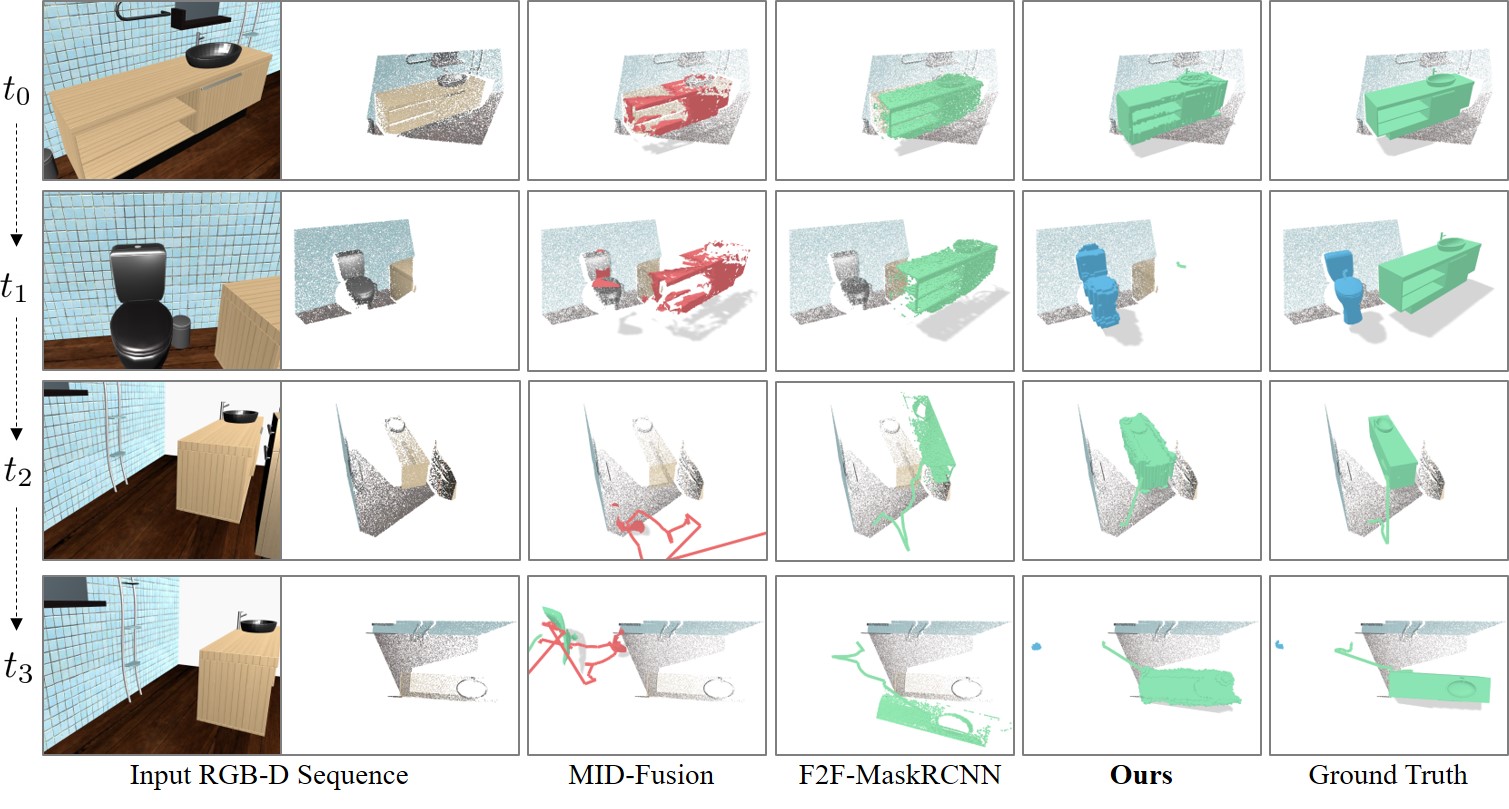}
	\caption{ 
	    Qualitative comparison to state of the art on \DYNSYNTH{} test sequences. 
	    Our approach predicting the complete object geometry maintains strong correspondence overlap even when objects or camera undergo stronger motions, resulting in notably more robust tracking that state-of-the-art approaches considering only the observed geometry.
    }
	\label{fig:comparison_dynsynth}
\end{figure*}

We evaluate our approach both quantitatively and qualitatively on synthetic RGB-D sequences of moving objects, as well as on real-world RGB-D data. 
We use a synthetic dataset, \DYNSYNTH{}, which contains $3,300$ RGB-D sequences of indoor scenes (2900/300/100 train/val/test), comprising $97,626$ frames.
We focus on detecting and tracking objects of $10$ class categories covering a variety of bedroom, living room, and bathroom furniture.
Each sequence contains camera trajectories and an object moving parallel to the ground, and ground truth object symmetries are provided.

As ground truth is available by nature of the synthetic data generation, we can train and fully evaluate our approach on \DYNSYNTH{}.
We also evaluate our object pose estimation on real-world, static RGB-D scans from the ScanNet data set~\cite{dai2017scannet} with ground truth object annotations provided by Scan2CAD~\cite{avetisyan2019scan2cad}.
We follow the official train/val/test split with Scan2CAD annotations with 944/149/100 scans, resulting in $114,000$ frames (sampled every 20th frame from the video sequences).

\paragraph{Evaluation metrics.}
To evaluate our dynamic object tracking, we adopt the Multiple Object Tracking Accuracy metric~\cite{bernardin2008evaluating}, which summarizes error from false positives, missed targets, and identity switches: 
\begin{equation}
    \textrm{MOTA} = 1 - \sum_t \frac{(m_t + fp_t + mme_t)}{\sum_t gt}
\end{equation}
where $m_t$, $fp_t$, $mme_t$ are number of misses, of false positives and of mismatches at time $t$.

A match is considered positive if its $\ell_2$ distance to ground truth center is less than $25$cm.
The state-of-the-art approaches that we evaluate predict only surface correspondences, so we establish their trajectories by shifting from the initial pose towards the ground truth center. 
We report the mean MOTA over all test sequences.

\paragraph{Comparison to state of the art.}
In Table~\ref{tab:mota_class}, we show that our approach to jointly complete and track objects provides significant improvement over state of the art on synthetic sequences from the \DYNSYNTH{} dataset.

We compare to MaskFusion~\cite{runz2018maskfusion}, a surfel-based approach for dense object tracking and reconstruction. MaskFusion's segmentation refinement step is unable to handle objects with non-convex surface or disconnected topology due to the self-occlusion and its weighted surfel tracking mechanism is not robust in the highly dynamic scenes (i.e. new information tends to be discarded).

We evaluate against MID-Fusion~\cite{xu2019mid}, a volumetric octree-based, dense tracking approach; MID-Fusion use volumetric representation to alleviate the low recall issue of its detection backend. However, it has a limited ability to align occluded objects with the existed models and associate proposals under fast object movement such as the qualitative examples in Figure~\ref{fig:comparison_real} and~\ref{fig:comparison_dynsynth}..

Additionally, we provide a baseline approach which performs frame-to-frame tracking for each object using the Iterative Closest Point algorithm~\cite{besl1992method,rusinkiewicz2001efficient}, given 2D detection provided by Mask R-CNN~\cite{he2017mask} trained on \DYNSYNTH{} (\emph{F2F-MaskRCNN}).  Searching correspondences between frames performs better under fast motion but it cannot resolve the weak geometry signals issue~\cite{gelfand2003stableicp} of the occluded objects such as the chair objects in Figure~\ref{fig:comparison_real}.

In contrast to these approaches which only reason based on the observed geometry from each view, our approach to infer the complete object geometry enables more robust and accurate object tracking. 

\paragraph{Does object completion help tracking?}
We analyze the effect of our object completion on both dynamic object tracking performance as well as pose estimation in single frames.
In Table~\ref{tab:mota_class}, we evaluate our approach on variants without object completion (\emph{no compl.}) or no correspondence-based object association (\emph{no corr.}); 
When matching is fully based on 3D bounding box overlap, we notice a small improvement of tracking performance of the variant with completion \emph{(no corr.)} over no completion (\emph{no corr., no compl.}) of 1.6\% mean MOTA. 
When association is based on canonical correspondences without using object completion (\emph{no compl.}), we observe a performance gain of 5\% mean MOTA. Utilizing object completion with canonical correspondences matching further improves the tracking performance by 6.7\% mean MOTA and achieves best results (42.3\% mean MOTA).

\begin{figure}
	\centering
	\includegraphics[width=\linewidth]{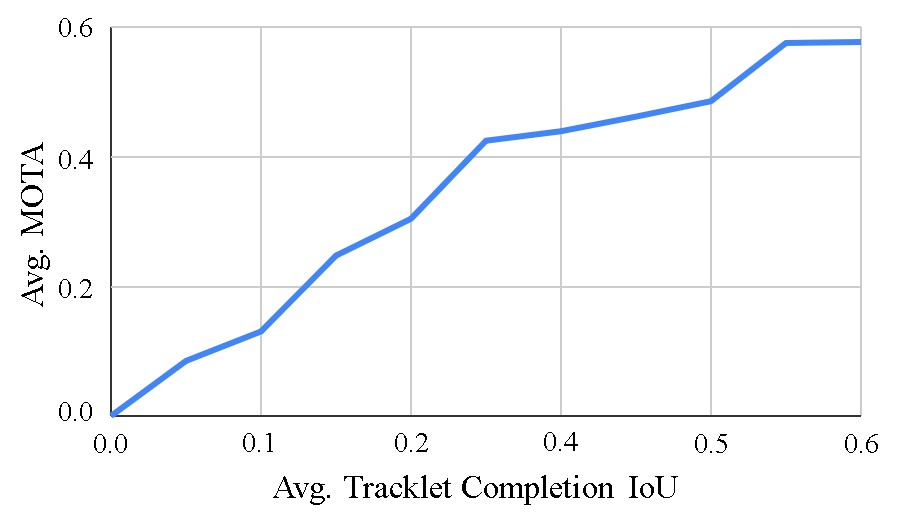}

	\caption{ 
	   Average tracking performance against average completion performance evaluated on \DYNSYNTH{} using our method. Better completion performance results in improved tracking, as correspondences can be more robustly established. 
    }
	\label{fig:mota_iou}
\end{figure}

Additionally, we show that our joint object completion and tracking improves on pose estimation for each object in individual frames.
Tables~\ref{tab:pose_eval_suncg} and \ref{tab:pose_eval_scannet} evaluate our approach with and without object completion on RGB-D frames from synthetic \DYNSYNTH{} data and real-world ScanNet~\cite{dai2017scannet} data, respectively.
We similarly find that for object pose estimation, inferring the complete underlying geometric structure of the objects provides more accurate object pose estimation.
Furthermore, we analyse in Figure \ref{fig:mota_iou} the tracking performance of our method with respect to the average completion performance on predicted tracklets. We observe that better completion also results in improved tracking, by facilitating correspondence in originally unobserved regions.

\paragraph{Real-world dynamic RGB-D sequences.}
In addition to the static RGB-D sequences of ScanNet~\cite{dai2017scannet}, we apply our approach to eight real-world dynamic RGB-D sequences which we captured with a Structure Sensor\footnote{https://structure.io/} mounted to an iPad.
In this scenario, we lack ground truth annotations, so we pre-train our model on \DYNSYNTH{} and fine-tune on ScanNet+Scan2CAD data.
Qualitative results are shown in  Figure~\ref{fig:comparison_real}; our approach finds persistent correspondences on the predicted complete object geometry, enabling robust object pose estimation and surface tracking.

\begin{table}[h]
\begin{center}
\begin{tabular}{l|c|c}
\hline
DynSynth   & Med rot. err. & \multicolumn{1}{l}{Med transl. err.} \\ \hline 
Ours (no compl.) & 7.4$^{\circ}$                  & 15.4cm                                         \\ \hline
Ours & \textbf{5.7$^{\circ}$}                   & \textbf{12.3cm}                                         \\ \hline
\end{tabular}
\end{center}
\vspace{0.1cm}
\caption{Evaluation of object pose estimation on individual RGB-D frames from \DYNSYNTH{}.
    Predicting the underlying geometry of each object enables more accurate object pose estimation in each frame.
    }
\label{tab:pose_eval_suncg}
\end{table}

\begin{table}[h]
\begin{center}
\begin{tabular}{l|c|c}
\hline
ScanNet+Scan2CAD     & Med rot. err. & \multicolumn{1}{l}{Med transl. err.}  \\ \hline
Ours (no compl.) &       16.6$^{\circ}$                  &  22.0cm                                             \\ \hline
Ours  &       \textbf{13.3$^{\circ}$}                &       \textbf{18.3cm  }                                      \\ \hline
\end{tabular}
\end{center}
\caption{Evaluation of object pose estimation on individual RGB-D frames from ScanNet~\cite{dai2017scannet}.
    Understanding the complete object geometry enables more reliable correspondence prediction for object pose estimation.}
\label{tab:pose_eval_scannet}
\end{table}

\begin{comment}
\begin{table*}[h]
\setlength{\tabcolsep}{4pt}
\centering
\begin{tabular}{l|ccccccccccc|c}
\toprule
Completion &  bathtub & bed   & bookshelf & cabinet & chair  & counter   & desk  & sink  & sofa  & table & toilet & mAP  \\ \midrule
\DYNSYNTH{}   &     &    &      &    &   &     &   &   &   &  &   &  \\
ScanNet  &    &  &       &  &  &     &  &  &  &   &    &  \\

\end{tabular}
\vspace{0.1cm}
\caption{
    }
	\label{tab:completion_map}
\end{table*}
\end{comment}

\section{Conclusion}

We introduce an approach for multi-object tracking in RGB-D sequences by learning to jointly infer the complete underlying geometric structure for each object as well as its dense correspondence mapping for pose estimation and tracking.
By predicting object geometry in unobserved regions, we can obtain correspondences that are more reliably persist across a sequence, producing more robust and accurate object tracking under various camera changes and occlusion patterns.
We believe that this provides significant promise in integration with a full reconstruction pipeline to perform live tracking and reconstruction of dynamic scenes towards object-based perception of environments.

\section*{Acknowledgments}

This work was supported by the ZD.B (Zentrum Digitalisierung.Bayern), a TUM-IAS Rudolf M\"o{\ss}bauer Fellowship, the ERC Starting Grant Scan2CAD (804724), and the German Research Foundation (DFG) Grant Making Machine Learning on Static and Dynamic 3D Data Practical. Yu-Shiang was partially supported by gifts from Adobe and Autodesk. 

\clearpage
{\small
\bibliographystyle{ieee_fullname}
\bibliography{egbib}
}
 \clearpage

\begin{appendix}
 \section*{Appendix}
In this appendix, we provide further details about our proposed method.
Specifically, we describe the network architectures in detail in Section~\ref{sec:network_details} and provide more quantitative results in Section~\ref{sec:addition_quant_eval}.

\section{Additional Quantitative Evaluation}
 \label{sec:addition_quant_eval}

We provide per-frame model performance on real-world ScanNet+Scan2CAD and the synthetic dataset \DYNSYNTH{}. 
In Table \ref{tab:detection_map}, we show class-wise detection results evaluated as mean average precision at a 3D IoU of 0.5 (mAP@0.5). The per-frame completion performance is evaluated in Table \ref{tab:completion_map} using a mean average precision metric with mesh IoU threshold of 0.25 (mAP@0.25).

\begin{table*}[hb]
\setlength{\tabcolsep}{4pt}
\centering

\begin{tabular}{l|cccccccccc|c}
\hline
       & bathtub & bed  & bookshelf & cabinet & chair & desk & sink & sofa & table & toilet & mAP  \\ \hline
\DYNSYNTH{}       & 49.3    & 38.4 & 12.5      & 6.3     & 44.1  & 46.8 & 27.6 & 32.3 & 38.4  & 63.1   & 35.8 \\
ScanNet+Scan2CAD &    38.7   &  -  &  12.9     & 4.6      & 41.2 &  -   & -     &  26.4 & 29.2  & -  &    25.6 \\      
\end{tabular}
\caption{
    3D Detection results on \DYNSYNTH{} and ScanNet with Scan2CAD targets at mAP@0.5.  
    }
	\label{tab:detection_map}
\vspace{0.5cm}
\end{table*}

\begin{table*}[hb]
\setlength{\tabcolsep}{4pt}
\centering
\begin{tabular}{l|cccccccccc|c}
\hline
      & bathtub & bed  & bookshelf & cabinet & chair & desk & sink & sofa & table & toilet & mAP  \\ \hline
\DYNSYNTH{}       & 34.8 & 23.6  & 12.7 & 11.4& 38.4 & 34.1 & 32.2& 41.1& 29.9& 52.6&31.1\\
ScanNet+Scan2CAD &   20.4 & - & 8.6 & 12.7 & 24.4 & - & - & 23.9 & 12.2 & - & 17.1\\      
\end{tabular}
\caption{
    Instance Completion results on \DYNSYNTH{} and ScanNet with Scan2CAD targets at mAP@0.25.  
    }
	\label{tab:completion_map}
\vspace{1cm}
\end{table*}

\section{Network Details}
 \label{sec:network_details}

\begin{figure*}
	\centering
	\includegraphics[width=\linewidth]{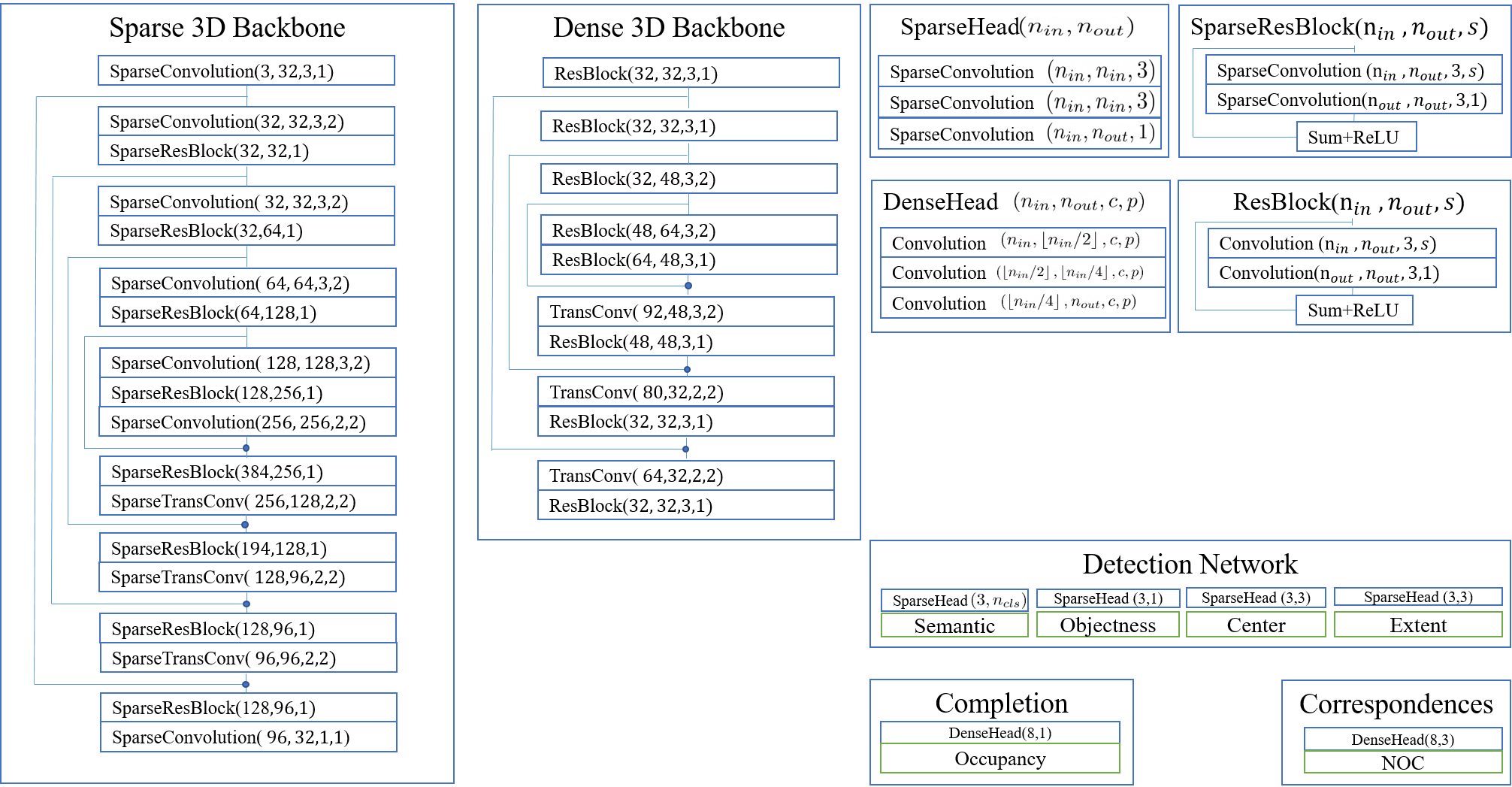}

	\caption{
	    Network architecture specification for our approach. Dots indicate concatenation, outputs are highlighted in green.
    }
	\label{fig:network_arc}
\end{figure*}

We detail the architecture of our network in Figure \ref{fig:network_arc}.  We provide the convolution parameters as (n\_in, n\_out, kernel\_size, stride, padding), where stride and padding default to 1 and 0, respectively.  Each convolution (except the last) is followed by batch normalization and a ReLU. 
 
 \end{appendix}

\end{document}